# Towards Transparency of TD-RL Robotic Systems with a Human Teacher

Marco Matarese
marco.matarese@studenti.unina.it
DIETI, University of Naples Federico II
Napoli, Italy

Silvia Rossi
silvia.rossi@unina.it
DIETI, University of Naples Federico II
Napoli, Italy

Alessandra Sciutti
alessandra.sciutti@iit.it
CONTACT, Istituto Italiano di Tecnologia
Genova, Italy

Francesco Rea
francesco.rea@iit.it
RBCS, Istituto Italiano di Tecnologia
Genova, Italy

## ABSTRACT

The high request for autonomous and flexible HRI implies the necessity of deploying Machine Learning (ML) mechanisms in the robot control. Indeed, the use of ML techniques, such as Reinforcement Learning (RL), makes the robot behaviour, during the learning process, not transparent to the observing user. In this work, we proposed an emotional model to improve the transparency in RL tasks for human-robot collaborative scenarios. The architecture we propose supports the RL algorithm with an emotional model able to both receive human feedback and exhibit emotional responses based on the learning process. The model is entirely based on the Temporal Difference (TD) error. The architecture was tested in an isolated laboratory with a simple setup. The results highlight that showing its internal state through an emotional response is enough to make a robot transparent to its human teacher. People also prefer to interact with a responsive robot because they are used to understand their intentions via emotions and social signals.

## 1 INTRODUCTION

The more robots become autonomous and flexible, the more their behaviour need to be transparent to effectively collaborate with a human user. This necessity becomes stronger when dealing with machine learning algorithms controlling the robot's behaviour. RL agents make errors during their learning process, not just because they are designed to do so, but because errors and exploration are intrinsically part of RL. So, we need robots that can understand us but at the same time can be easily understood and anticipated by us. To improve this mutual understanding we need to introduce transparency into the robots' behaviours.

Moreover, Broekens and Chetouani, in their position paper [3], argue that the lack of transparency in the robot behaviour may have a direct impact on robot learning. Such transparency can be obtained through the use of nonverbal cues. Humans and other animals use such nonverbal signals to express their internal state. Among them, emotion expression is one of more important because it is language and species independent. Indeed, they argue that simulation of emotions could be used to make learning robots more transparent to their human users and co-workers.

In this paper, we focused on RL, a powerful learning method that, due to its try and error behaviour, intrinsically lacks in transparency. We designed a model based on RL starting from a TD-RL Theory of Emotion [2]. Our model enables an agent to select the appropriate emotion to communicate its internal state and detect and interpret human feedback in terms of emotions to be used as learning signals. To test the proposed approach, we designed an experiment to evaluate how an emotional response based on the robot learning process can make it more transparent to humans while it is performing a RL task.

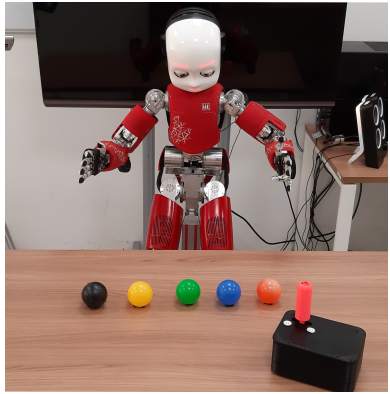

Figure 1: Experimental setup.

## 2 METHODS

Our experimental hypothesis is that the robot emotional response is sufficient to make the robot behaviour more transparent to the human teacher. We defined the robot response through three non-verbal communicative channels: the (un)certainty of the movement, the facial expression, and the gaze. Each channel is used to display a particular *feel*: how much the robot is confident about actions, the degree of satisfaction for the actions' effects, and an anticipatory signal for the next chosen action, respectively. The robot multi-modal behaviour is selected according to the learning process and, in particular, on the TD error. It is a local estimate of the learning trend giving us information about the goodness of the last performed action in comparison to the previous one.

For our experiments, we chose a very simple RL task that could be used in an HRI scenario: the robot had to learn a specific, defined a priori by the investigator, sequence of objects. The experimental setup is shown in Figure 1: the robot iCub was placed on a fixed metal support in front of a table, on which there were five balls aligned in fixed positions and a joystick; on the other side of the table, there was the participants' chair. We had 23 participants, their average age was $\mu_{age} \simeq 27$ with a standard deviation of $\sigma_{age} \simeq 8$. The experiments consisted of two sessions (mechanical and human-like): in both, the robot tried to learn the right sequence of balls taking into account the human feedback, but in one session the robot showed a mechanical behaviour and in the other one it showed the emotional response. The participants gave their feedback through the joystick. The feedback was continuous and it could be negative, neutral or positive. During the mechanical sessions, the robot just pointed the balls to indicate the chosen sequences. In the human-like behaviour, we had both certain and uncertain robot movements. When the robot showed uncertain movements it, before the proper pointing, moved its hand in midair to appear thoughtful and uncertain. Then, it moved its gaze stopping it, in the end, towards the chosen object. Starting from this uncertain pose, the robot movement continued by pointing the chosen object. On the other hand, if the robot was certain, it first used an anticipatory gaze signal and then it pointed the chosen object. That is why we have just one vertical line in Figure 2.b, two vertical lines in 2.c, and no lines at all in Figure 2.a. The vertical lines indicate when a robot movement ends and another one begins. Face emotional reaction were modelled concerning the TD error (positive, negative or neutral). We submitted the participants the Godspeed questionnaire [1] and the Mind Attribution Test [4] to discover any perceived differences between the robot behaviours.

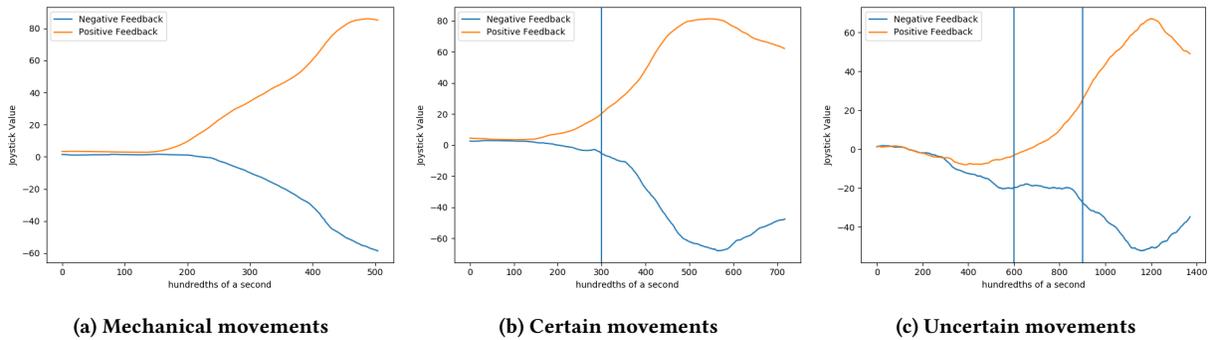

**Figure 2: Average of human feedback given through the joystick. The maximum was +100, the minimum was -100.**

## 3 RESULTS

Figure 2 shows the average participants' feedback in both kinds of sessions. Figure 2.a concerns the mechanical sessions, the others concern the human-like ones. As we can see from the picture, during the mechanical sessions, the participants' feedback was concentrated at the end of the robot action: in average, they begin in the middle of the robot pointing. During the human-like sessions instead, the participants started giving their feedback during the robot anticipatory gaze so, when the robot pointing started, they were already giving relevant feedback. We can claim that the human-like behaviour allowed people to anticipate their feedback. We can also see that during the uncertain robot movement people gave negative feedback in both cases: very little for right actions and stronger for wrong actions. This type of action was not pointing or a signal communicating intentions. It seems that people wanted to communicate something to the robot noticing that it was in need of help. From the plots is clear that the anticipatory gaze was well perceived by the participants and that it had a fundamental role during the interaction: in both certain and uncertain behaviour, the joystick use drastically changed during this signal.

We did paired t-tests looking for differences between the average of questionnaires replies relating to the two kinds of sessions. We used a confidence interval of 95%. We found significant differences in replies regarding the robot ability to feel pleasure ($t(22) = -3.49$, $p = 0.0020$) and joy ($t(22) = -4.54$, $p = 0.0001$), and in those about the robot Anthropomorphism ($t(22) = -4.57$, $p = 0.0001$), Animacy ($t(22) = -5.86, p < 0.0001$) and Likeability ($t(22) = -2.22$, $p = 0.0367$). For each questionnaire, we registered that all the participants' replies regarding human-like sessions were higher than those regarding the mechanical sessions.

## 4 CONCLUSIONS

Regarding the questionnaires' results, we can claim that the participants well perceived the differences between the two robot behaviours and that they appreciated more the human-like one. It is clear the main role played by the gaze: in all human-like scenarios, the feedback's trend was quite the same. Only the feedback given to uncertain movements followed by wrong actions seem to stop during the robot gaze change as if the participants were uncertain too, and then continue as expected. In all other human-like scenarios, the user feedback started decisively during the gaze, so before the pointing movement's would begin. So, we can claim that the anticipatory gaze was enough to understand the robot's intentions.